\documentclass[letterpaper, 10pt, journal, twoside]{IEEEtran}

\usepackage[utf8]{inputenc}
\usepackage{amssymb,amsfonts,amsmath,amscd}
\usepackage{bm}
\usepackage[pdftex]{graphicx}
\usepackage{url}
\usepackage{cite}
\usepackage[usenames,dvipsnames]{color}
\usepackage{subcaption}
\usepackage{caption}
\usepackage{accents}
\usepackage{tabularx}
\usepackage{tikz}
\usepackage{stackengine}

\usepackage{enumitem}

\usepackage{mathtools}
\usepackage[english]{babel}
\usepackage{siunitx}
\usepackage[export]{adjustbox}

\usepackage{microtype}

\usepackage{booktabs}
\usepackage{makecell}
\usepackage{multirow}

\definecolor{pltred}{rgb}{0.839, 0.153, 0.157}

\DeclareMathOperator*{\argmin}{argmin}

\DeclareMathOperator{\diag}{\mathrm{diag}}

\newcommand{\R}{\mathbb{R}}

\newcommand{\ubar}[1]{\underaccent{\bar}{#1}}

\newcommand{\videourl}{http://tiny.cc/keep-it-upright}
\newcommand{\codeurl}{https://github.com/utiasDSL/upright}

\title{Keep it Upright: Model Predictive Control for Nonprehensile Object Transportation with Obstacle Avoidance on a Mobile Manipulator}
\author{Adam Heins and Angela P. Schoellig%
\thanks{This is the long version of an article published in IEEE Robotics and Automation Letters. Digital Object Identifier (DOI): 10.1109/LRA.2023.3324520}%
\thanks{This work was supported by the Natural Sciences and Engineering Research Council of Canada and the Canadian Institute for Advanced Research.}%
  \thanks{The authors are with the Learning Systems and Robotics Lab (www.learnsyslab.org) at the Technical University of Munich, Germany, and the University of Toronto Institute for Aerospace Studies, Canada. They are also affiliated with the University of Toronto Robotics Institute, the Munich Institute of Robotics and Machine Intelligence (MIRMI), and the Vector Institute for Artificial Intelligence. E-mail: adam.heins@robotics.utias.utoronto.ca, angela.schoellig@tum.de}%
  \thanks{\copyright\ 2023 IEEE.  Personal use of this material is permitted.  Permission from IEEE must be obtained for all other uses, in any current or future media, including reprinting/republishing this material for advertising or promotional purposes, creating new collective works, for resale or redistribution to servers or lists, or reuse of any copyrighted component of this work in other works.}%
}

\begin{document}

\maketitle

\begin{abstract}
  We consider a nonprehensile manipulation task in which a mobile manipulator
  must balance objects on its end effector without grasping them---known as the \emph{waiter's
  problem}---and move to a desired location while avoiding static and dynamic
  obstacles. In contrast to existing approaches, our focus is on fast online
  planning in response to new and changing environments. Our main contribution
  is a whole-body constrained model predictive controller (MPC) for a mobile
  manipulator that balances objects and avoids collisions. Furthermore, we
  propose planning using the minimum statically-feasible friction coefficients,
  which provides robustness to frictional uncertainty and other force
  disturbances while also substantially reducing the compute time required to
  update the MPC policy. Simulations and hardware experiments on a
  velocity-controlled mobile manipulator with up to seven balanced objects,
  stacked objects, and various obstacles show that our approach can handle
  a variety of conditions that have not been previously demonstrated, with end
  effector speeds and accelerations up to~2.0~m/s
  and~7.9~m/s\textsuperscript{2}, respectively. Notably, we demonstrate a
  projectile avoidance task in which the robot avoids a thrown ball while
  balancing a tall bottle.
\end{abstract}

\section{Introduction}

\begin{figure}[t]
  \centering
  \includegraphics[width=\columnwidth,trim={0 0mm 0 0mm},clip]{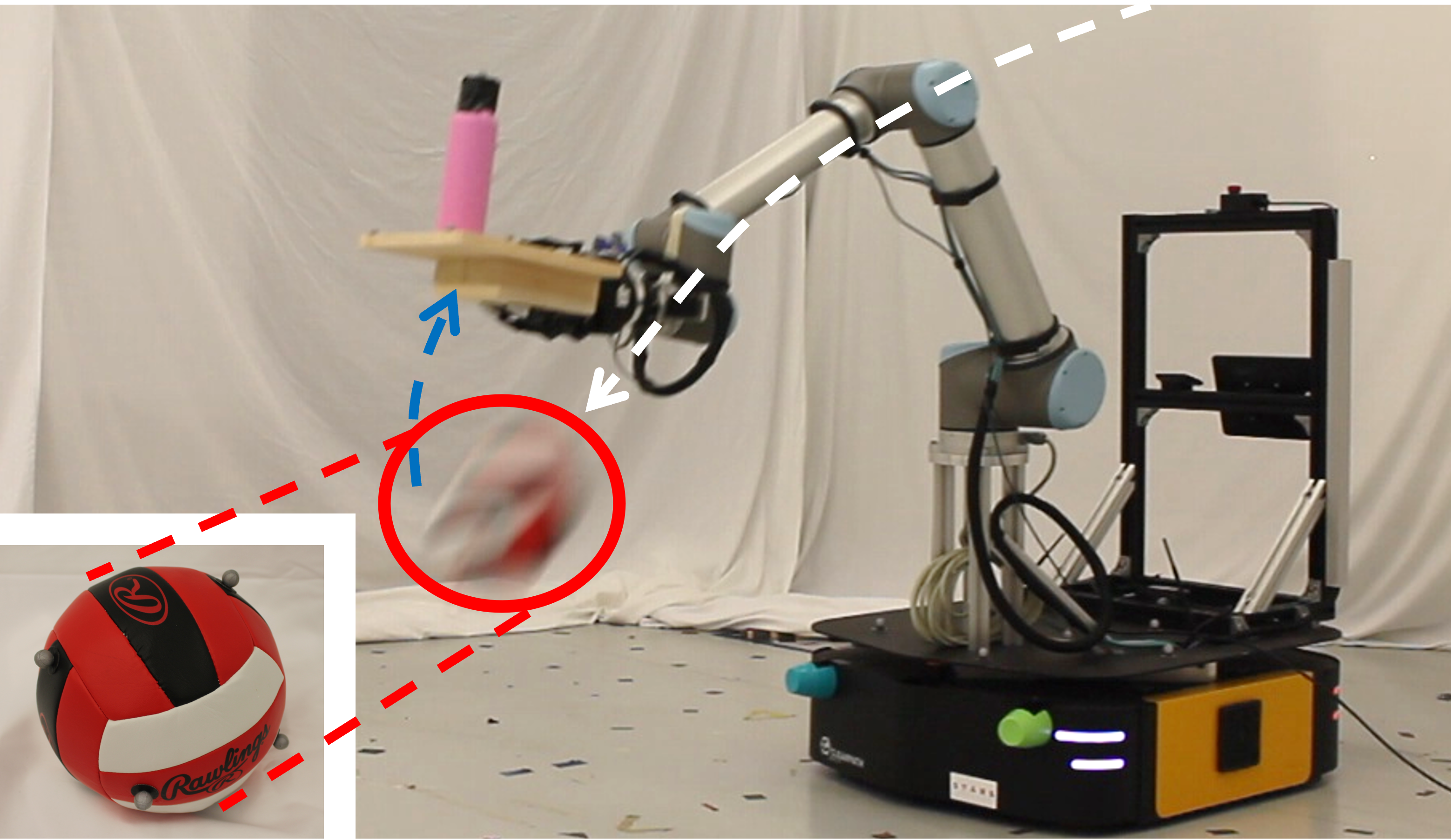}
    \caption{Our mobile manipulator balancing a pink bottle while avoiding a thrown
      volleyball (ball circled in red with approximate trajectory in white;
      approximate end effector trajectory in blue). The controller has less
      than~\SI{0.75}{s} between first observing the ball and a potential
      collision. A video of our experiments is available at
      \textbf{\texttt{\scriptsize \videourl}}.}
  \label{fig:eyecandy}
\end{figure}

We consider the nonprehensile object transportation task known as the
\emph{waiter's problem}~\cite{flores2013time}, which requires the robot to
transport objects from one location to another while keeping them balanced on a
tray at the end effector (EE), like a restaurant waiter. \emph{Nonprehensile}
manipulation~\cite{lynch1996nonprehensile} refers to the case when the
manipulated objects are subject only to unilateral
constraints~\cite{ruggiero2018nonprehensile} and thus retain some degrees of
freedom (DOFs); that is, they are not fully grasped. In contrast to
\emph{prehensile} manipulation, a nonprehensile approach allows the robot to
carry many objects at once with a simple, non-articulated EE (e.g., a tray; see
Fig.~\ref{fig:eyecandy} and~\ref{fig:experiment_setup}). Furthermore, a
nonprehensile approach skips the potentially slow grasping and ungrasping
processes, and can handle small or delicate objects which cannot be adequately
grasped~\cite{pham2017admissible}. Beyond food service, efficient object
transportation is useful across many industries, such as warehouse fulfillment
and manufacturing.

Specifically, we address the waiter's problem using a velocity-controlled
mobile manipulator. Mobile manipulators are capable of performing a wide
variety of tasks due to the combination of the large workspace of a mobile base
and the manipulation capabilities of robotic arms.
We are particularly interested in having the mobile manipulator move and
react \emph{quickly}, whether to avoid obstacles or simply for efficiency.
However, a challenge of \emph{mobile} manipulation is that moving
across the ground causes vibration at the EE, which requires our object balancing
strategy to be robust to such disturbances.

The goal of this work is to develop a controller for a mobile manipulator to
quickly transport objects to a desired location without dropping them
or colliding with any static or dynamic obstacles, the trajectories of which
may not be known a priori. Objects are held on a tray at the EE under
frictional contact (i.e., without the use of grasping or adhesive), and they
should neither fall over nor slip off the tray. We assume that the geometry,
inertial properties, and initial poses of the objects are known, but we do not
assume that feedback of the objects' poses is available online. We assume the
robot is velocity-controlled and a kinematic model is available; its
dynamic model is not required.

This work makes the following contributions:
\begin{enumerate}
  \item \textbf{Control:} We propose the first whole-body model predictive
    controller (MPC) for a mobile manipulator solving the waiter's problem.
    Compared to existing MPC-based approaches to this problem, which have only been
    demonstrated on fixed-base arms, our controller optimizes the
    joint-space trajectory online directly from task-space objectives and
    constraints, without the use of a higher-level planning step.
    Furthermore, the controller uses the minimum statically-feasible friction
    coefficients, which provides robustness to frictional uncertainty,
    vibration, and other real-world disturbances. When the minimum statically-feasible friction coefficients are \emph{zero}, we show that the MPC problem
    can be solved more efficiently.
  \item \textbf{Experiments:} We present the first demonstrations of the
    waiter's problem with a real velocity-controlled mobile manipulator
    balancing up to seven objects; balancing an assembly of stacked objects;
    and avoiding static and dynamics obstacles, including a thrown volleyball
    (see Fig.~\ref{fig:eyecandy}). The EE achieves speeds and accelerations up
    to~\SI{2.0}{m/s} and~\SI{7.9}{m/s\squared}, respectively.
  \item \textbf{Code:} Our code is available as an open-source library at
    \texttt{\small \codeurl}.
\end{enumerate}

After discussing related work in Sec.~\ref{sec:related} and background
information in Sec.~\ref{sec:model} and~\ref{sec:bal}, we present our robust
balancing constraints in Sec.~\ref{sec:robust} and our controller
in~Sec.~\ref{sec:mpc}. Simulations and hardware experiments follow
in~Sec.~\ref{sec:sim} and~\ref{sec:exp}, and Sec.~\ref{sec:conclusion}
concludes the paper.

\section{Related Work}\label{sec:related}

Prior examples of robots directly inspired by waiters in a restaurant
include~\cite{maxwell1999alfred,cheong2016development,wan2020waiter}, but these
are mobile robots without manipulator arms. In contrast, a mobile
\emph{manipulator} has additional DOFs that provide redundancy and a larger
workspace, at the cost of requiring a larger and more computationally-demanding
control problem.

One approach for balancing objects is to use some manner of sensor
feedback to infer the object states. In~\cite{sprenger1998balancing}, a
manipulator performs the classic inverted pendulum task.
In~\cite{permadi2014balancing}, a controller is developed to stabilize a tray
based on data from an attached accelerometer and gyroscope.
In~\cite{garciaharo2018balance}, an object is balanced on a tray by a humanoid
robot based on force-torque measurements from the robot's wrists.
While the focus of~\cite{garciaharo2018balance} is correcting for an
object's loss of balance, we focus on generating fast motions that
\emph{maintain} open-loop balance without object feedback.

A two-dimensional version of the waiter's problem is addressed
in~\cite{dang2004active}, in which a parallel manipulator is mounted on a
mobile robot to compensate for the sensed acceleration of the system. The
manipulator is controlled to act like a pendulum to minimize the tangential
forces acting on a transported object. Simulation of pendular motion has also
been used for the slosh-free transport of
liquids~\cite{moriello2018manipulating,muchacho2022a}, though these works focus
on imposing particular dynamics on the EE rather than directly constraining
its motion. EE acceleration constraints are imposed
in~\cite{ichnowski2022gomp} to avoid dropping grasped objects or spilling
liquids, but nonprehensile object transportation is not addressed.

The waiter's problem has also been addressed using offline motion planning.
Time-optimal path planning (TOPP) approaches minimize the time required to
traverse a provided path subject to the constraint that the transported objects
remain balanced. In~\cite{csorvasi2017near}, convex programming is used to
solve the TOPP problem. In~\cite{luo2017robust}, a robust time-scaling approach
is used to handle confidence bounds on model parameters like friction, which is
combined with iterative learning to learn the bounds. Other planning-based
approaches do not assume a path is provided. A kinodynamic RRT-based planner is
applied to the nonprehensile transportation task in~\cite{pham2017admissible},
which demonstrates solving a task where no quasistatic solution exists. An
optimization-based planner is applied to the task in~\cite{flores2013time}. In
contrast to these offline planning approaches, our method runs online to react
quickly to changes in the environment.

In~\cite{selvaggio2022a} and~\cite{subburaman2023a}, a reactive controller
automatically regulates the commanded motion to ensure the object remains
balanced. A similar approach is applied to legged robots
in~\cite{morlando2022nonprehensile}, where the desired trajectory is generated
by a spline-based planner. This is one of the only works to use a full mobile
manipulator (a quadruped) for the waiter's problem, but it is demonstrated only
in simulation and does not consider dynamic obstacles. To our knowledge, the
only physical mobile manipulator experiments for the waiter's problem have been
performed on a humanoid in~\cite{haro2019object}, but similar
to~\cite{garciaharo2018balance} they focus on the detection and
rejection of disturbances to the object's stability rather than fast object
transportation.

Finally, like us, some recent works use MPC to address the waiter's problem.
In~\cite{zhou2022topp}, a dual-arm approach is proposed in which a time-optimal
trajectory is planned and MPC is used to compute the applied wrench required to
realize the object's trajectory. Another MPC approach is described
in~\cite{selvaggio2023non}, which is designed to track a manipulator's
joint-space reference trajectory. In contrast, our MPC approach optimizes the
joint-space trajectory online while considering task-space objectives and
constraints, which allows us to respond quickly to changes in the environment
like dynamic obstacles, and we also show how reducing the friction coefficients
in the controller constraints can provide robustness and computational savings.

\section{System Model}\label{sec:model}

We start with the models of the robot and balanced objects.

\subsection{Robot Model}

We consider a velocity-controlled mobile manipulator with
state~$\bm{x}=[\bm{q}^T,\bm{v}^T,\dot{\bm{v}}^T]^T$, where~$\bm{q}$ is the
generalized position, which includes the planar pose of the mobile base and the
arm's joint angles, and~$\bm{v}$ is the generalized velocity. We include
acceleration in the state and take the input~$\bm{u}$ to be jerk, which ensures a
continuous acceleration profile~\cite{selvaggio2023non}. The input is
double-integrated to obtain the velocity commands sent to the actual robot. We
require only a kinematic model, which we represent generically as
\begin{equation*}
  \dot{\bm{x}} = \bm{a}(\bm{x}) + \bm{B}(\bm{x})\bm{u},
\end{equation*}
with~$\bm{a}(\bm{x})\in\R^{\mathrm{dim}(\bm{x})}$ and~$\bm{B}(\bm{x})\in\R^{\mathrm{dim}(\bm{x})\times\mathrm{dim}(\bm{u})}$.

\subsection{Object Model}

We model each object~$\mathcal{O}$ as a rigid body subject to the Newton-Euler
equations
\begin{equation}\label{eq:obj_dynamics}
  \bm{w}_{\mathrm{C}} + \bm{w}_{\mathrm{GI}} = \bm{0},
\end{equation}
where~$\bm{w}_{\mathrm{C}}$ is the total contact wrench
and~$\bm{w}_{\mathrm{GI}}$ is the gravito-inertial wrench, expressed in the
body frame as
\begin{equation*}
  \bm{w}_{\mathrm{GI}} \triangleq \begin{bmatrix} \bm{f}_{\mathrm{GI}} \\ \bm{\tau}_{\mathrm{GI}} \end{bmatrix} = -\begin{bmatrix}
    m(\dot{\bm{v}}_{o} - \bm{R}_o\bm{g}) \\
    \bm{J}\dot{\bm{\omega}}_o + \bm{\omega}_o\times\bm{J}\bm{\omega}_o
  \end{bmatrix},
\end{equation*}
where~$\bm{f}_{\mathrm{GI}}$ and $\bm{\tau}_{\mathrm{GI}}$ are the
gravito-inertial force and torque, $m$ is the object's mass, $\bm{v}_o$
and~$\bm{\omega}_o$ are the body-frame linear and angular velocity of the
object's CoM, $\bm{g}$ is the gravitational acceleration, and~$\bm{J}$ is the
object's inertia matrix taken about the CoM. The rotation matrix~$\bm{R}_o$
represents the object's orientation with respect to the world and is used to
rotate gravity into the body frame.

\section{Balancing Constraints}\label{sec:bal}

To control the interaction between the EE and balanced objects in the most
general case, we would need to reason about the hybrid dynamics resulting from
different contact modes (sticking, sliding, no contact, etc.). Instead, our
approach is to enforce constraints that keep the system in a single mode
(sticking); that is, we constrain the robot's motion so that the balanced objects
do not move with respect to the EE. This is known as a \emph{dynamic
grasp}~\cite{lynch1996nonprehensile}. Now assuming the object is in the
sticking mode, we define the object's orientation as~$\bm{R}_o=\bm{R}_e$, such
that it is aligned with the EE's orientation~$\bm{R}_e$. Furthermore, we
have~$\bm{v}_{o}=\bm{v}_e+\bm{\omega}_e\times\bm{c}$
and~$\bm{\omega}_{o}=\bm{\omega}_e$, where~$\bm{v}_e$ and~$\bm{\omega}_e$ are
the EE's linear and angular velocity in the body frame and~$\bm{c}$ is the
position of the object's CoM with respect to the EE. Thus we can write the
object's gravito-inertial wrench as
\begin{equation}\label{eq:GI_in_terms_of_ee}
  \bm{w}_{\mathrm{GI}} = -\begin{bmatrix}
    m(\dot{\bm{v}}_e - \bm{R}_e\bm{g}) + m(\dot{\bm{\omega}}^\times_e+\bm{\omega}^{\times}_e\bm{\omega}^{\times}_e)\bm{c} \\
    \bm{J}\dot{\bm{\omega}}_e + \bm{\omega}_e^\times\bm{J}\bm{\omega}_e
  \end{bmatrix},
\end{equation}
where~$(\cdot)^{\times}$ converts a vector to a skew-symmetric matrix such
that~$\bm{a}^{\times}\bm{b}=\bm{a}\times\bm{b}$ for any~$\bm{a},\bm{b}\in\R^3$. We
assume that the inertial parameters~$m$, $\bm{c}$, and~$\bm{J}$
are known. Let us group the remaining variables, along with the EE
position~$\bm{r}_e$, into the
tuple~$\bm{e}=(\bm{R}_e,\bm{r}_e,\bm{\varpi}_e,\dot{\bm{\varpi}}_e)$,
where~$\bm{\varpi}_e=[\bm{v}_e^T,\bm{\omega}_e^T]^T$ is the EE's generalized
velocity, which we refer to as the EE state. We can compute~$\bm{e}$ from the
robot state~$\bm{x}$ via forward kinematics, in which case we may explicitly
write~$\bm{e}(\bm{x})$. As can be seen
in~\eqref{eq:GI_in_terms_of_ee}, the object's motion is completely determined
by~$\bm{e}$ when sticking; the remainder of this section describes
the constraints required to maintain the sticking mode. We do not use online
feedback of the object state---given the initial object poses with respect to
the EE, the controller generates trajectories to keep those poses constant in
an open-loop manner.

\begin{figure}[t]
  \centering
    \includegraphics[width=0.6\columnwidth]{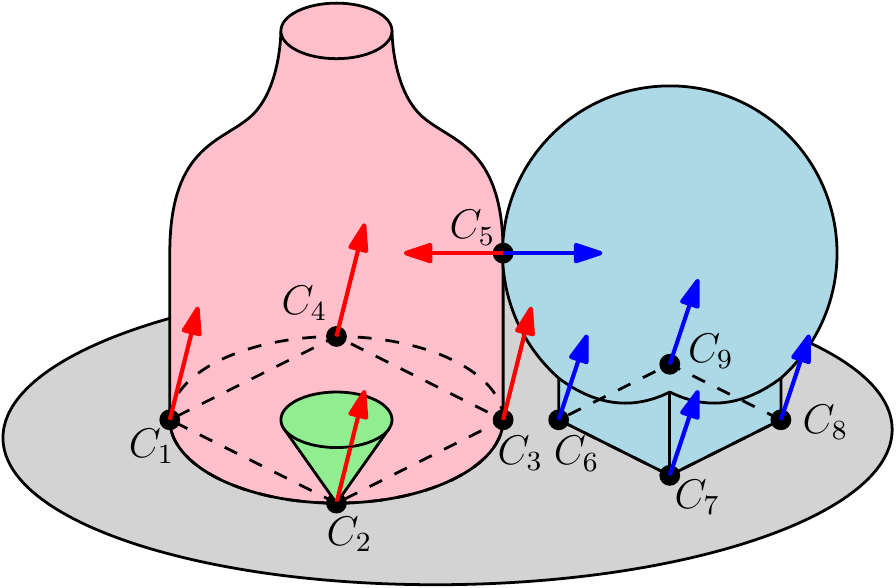}
    \caption{A bottle (red) and globe (blue) balanced on a tray. This
      arrangement has a total of $N=9$ contact points (black dots), with each
      object having~$n=5$ ($C_5$ is shared). Contact forces (arrows) at each
      contact point must belong to their friction cones (one shown in green).
      The circular contact patch of the bottle is approximated by a
      quadrilateral. The contact force acting on each object at the shared
      contact point~$C_5$ must be equal and opposite. If~$\mu_i=0$, the friction
      cone at~$C_i$ collapses to the line along the normal~$\hat{\bm{n}}_i$.}
  \label{fig:3d_diagram}
\end{figure}

A general approach for ensuring an object sticks to the EE can be obtained by
including all contact forces directly into the optimal control problem and
constraining the solution to be consistent with the desired (sticking)
dynamics, which has been previously applied to the waiter's problem in,
e.g.,~\cite{selvaggio2022a} and~\cite{selvaggio2023non}.
Consider an arrangement of objects with~$N$ total contact
points~$\{C_i\}_{i\in\mathcal{I}}$ and corresponding contact
forces~$\{\bm{f}_i\}_{i\in\mathcal{I}}$, where~$\mathcal{I}=\{1,\dots,N\}$
(see Fig.~\ref{fig:3d_diagram}). By Coulomb's law, each~$\bm{f}_i$ must be inside its friction cone. We use an inner pyramidal
approximation of the friction cone
\begin{equation}\label{eq:friction_cone}
  \|\bm{f}^t_i\|_1 \leq \mu_i f^n_i,
\end{equation}
where~$f^{n}_i\triangleq\hat{\bm{n}}_i^T\bm{f}_i$ is the force along the
contact normal~$\hat{\bm{n}}_i$, $\bm{f}^t_i$ is the force tangent
to~$\hat{\bm{n}}_i$, and~$\mu_i$ is the friction coefficient.

The total contact wrench acting on an individual object is
\begin{equation}\label{eq:contact_wrench}
  \bm{w}_{\mathrm{C}} \triangleq \begin{bmatrix} \bm{f}_\mathrm{C} \\ \bm{\tau}_\mathrm{C} \end{bmatrix} = \sum_{j\in\mathcal{J}}\begin{bmatrix} \bm{f}_j \\ \bm{r}_j\times\bm{f}_j \end{bmatrix},
\end{equation}
where~$\bm{f}_{\mathrm{C}}$ and~$\bm{\tau}_{\mathrm{C}}$ are the total contact
force and torque,~$\mathcal{J}\subseteq\mathcal{I}$ is the subset of contact
indices for this particular object, and~$\bm{r}_j$ is the location of~$C_j$
with respect to the object's CoM. The object is successfully balanced for a
given~$\bm{e}$ if a set of contact forces can be found each
satisfying~\eqref{eq:friction_cone} and consistent
with~\eqref{eq:obj_dynamics}, \eqref{eq:GI_in_terms_of_ee},
and~\eqref{eq:contact_wrench}. We assume that contact patches can be
represented as polygons with a contact point at each vertex; as
in~\cite{selvaggio2022a} we always use four points with equal~$\mu$.

However, we need an extra constraint for each contact point shared between two objects
(as opposed to contact points between an object and the tray; again refer to
Fig.~\ref{fig:3d_diagram}): per Newton's third law, the
contact force acting on each object must be equal and opposite.
Let~$\mathcal{O}^a$ and~$\mathcal{O}^b$ be two objects in contact at some
point~$C_i$, and denote~$\bm{f}^{a}_i$ and~$\bm{f}^b_i$ the corresponding
contact forces acting on~$\mathcal{O}^a$ and~$\mathcal{O}^b$, respectively.
Then we have the constraint
\begin{equation}\label{eq:contact_force_consistency}
  \bm{f}^{a}_i = -\bm{f}^{b}_i.
\end{equation}
To lighten the notation going forward, we gather all contact forces into the
vector~$\bm{\xi}=[\bm{f}_1^T,\dots,\bm{f}_N^T]^T$, and write
\begin{equation*}
  (\bm{e},\bm{\xi})\in\mathcal{B}
\end{equation*}
to indicate that the EE state~$\bm{e}$ and contact forces~$\bm{\xi}$ together
satisfy the balancing
constraints~\eqref{eq:obj_dynamics}--\eqref{eq:contact_force_consistency} for
all objects.

\section{Robust Contact Force Constraints}\label{sec:robust}

\begin{figure}[t]
  \centering
    \includegraphics[width=0.8\columnwidth]{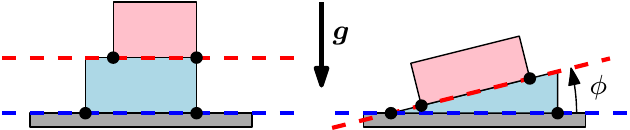}
    \caption{Planar view of two arrangements of objects, each with two objects
      balanced on a tray and a total of four contact points (black dots).
      \emph{Left:} the support planes (dashed lines) of each object
      are parallel, so the orientation shown is feasible in the presence of
      gravity with no friction forces (i.e., we can take~$\mu_i=0$ for
      all~$i\in\mathcal{I}$). \emph{Right:} the support planes are \emph{not}
      parallel, so some friction is \emph{always} required to balance this
      arrangement.}
  \label{fig:support_planes}
\end{figure}

The constraint~\eqref{eq:friction_cone} ensures all contact forces are inside their
respective friction cones. However, this assumes accurate knowledge of the
friction coefficients, and the constraint may also be violated by
unmodelled force disturbances like vibrations and air resistance. To improve the
controller's robustness, it is thus desirable for the tangential contact forces to
be small, keeping the forces away from the friction cone boundaries~\cite{selvaggio2022a}.
We propose to plan trajectories using the minimum statically-feasible values of
the friction coefficients; that is, the smallest coefficients for which there
exists an EE orientation~$\bm{R}_e$ and contact forces~$\bm{\xi}$ satisfying
the balancing constraints with zero EE velocity and acceleration. This ensures
that the controller can always converge to a stationary position.
Again considering an arbitrary arrangement of objects, we obtain the minimum
statically-feasible friction coefficients by solving the optimization
problem\footnote{It is not necessary to explicitly use a rotation matrix to
parameterize orientation in~\eqref{eq:minimum_mu}: any representation of
$SO(3)$ can be used. We use Euler angles. Note also that the balancing
constraints are independent of the EE's position, so we set~$\bm{r}_e=\bm{0}$.
See the appendix for more details.}
\begin{equation}\label{eq:minimum_mu}
  \begin{aligned}
    \argmin_{\bm{R}_e,\bm{\xi},\{\mu_i\}_{i\in\mathcal{I}}} &\quad \frac{1}{2}\sum_{i\in\mathcal{I}}\alpha_i\mu_i^2 \\
    \text{subject to} &\quad \mu_i \geq 0, \quad i\in\mathcal{I} \\
                      &\quad (\bm{e},\bm{\xi})\in\mathcal{B}, \\
                      &\quad \bm{e} = (\bm{R}_e,\bm{0},\bm{0},\bm{0}),
  \end{aligned}
\end{equation}
where~$\{\alpha_i\}_{i\in\mathcal{I}}$ are a set of weights. If we have nominal estimates
of the friction coefficients~$\{\bar{\mu}_i\}_{i\in\mathcal{I}}$, we set each
weight~$\alpha_i=1/\bar{\mu}_i$ to lower each coefficient proportionally;
otherwise we set~$\alpha_i=1$ for all~$i\in\mathcal{I}$.
In the common case when the support planes of each object are parallel to each
other (see Fig.~\ref{fig:support_planes}), the solution
to~\eqref{eq:minimum_mu} is simply~$\mu_i=0$ for all~$i\in\mathcal{I}$
with~$\bm{R}_e$ such that the support planes are orthogonal to gravity.
An example when the solution of~\eqref{eq:minimum_mu} is
not~$\mu_i=0$ for all~$i\in\mathcal{I}$ is discussed in
Sec.~\ref{sec:sim:nonparallel}. The problem~\eqref{eq:minimum_mu} need only be
solved once for a given arrangement of objects.

While choosing the minimum friction coefficients may at first appear overly
conservative, this approach has a number of benefits. First, it
removes the need for accurate friction coefficient estimates,
which requires time-consuming physical manipulation of the objects to
estimate. Second, \emph{mobile} manipulation can produce significant EE
vibration, requiring robust motions to ensure objects are balanced.
Third, in the common case when~$\mu_i=0$ for all~$i\in\mathcal{I}$, the
optimal control problem can be simplified as follows. In general we require
one contact force variable~$\bm{f}_i\in\R^3$ per contact point, each
constrained to satisfy~\eqref{eq:friction_cone}. However, when~$\mu_i=0$, we
can parameterize the force with a single scalar~$f_i\geq0$ such
that~$\bm{f}_i=f_i\hat{\bm{n}}_i$. This reduces the number of force decision
variables by two thirds and replaces~\eqref{eq:friction_cone} with a
simple bound, making the optimization problem faster to solve.

We solved~\eqref{eq:minimum_mu} assuming the EE was \emph{stationary}, since we
do not assume to know the full EE trajectories a priori. However, in general
it is not possible to accelerate multiple objects while
assuming \emph{zero} friction, even when there is a feasible stationary solution. To see
this, first consider a single object on a tray with its support plane
orthogonal to gravity and with~$\mu_i=0$ for
all~$i\in\mathcal{I}$. From~\eqref{eq:contact_wrench} we
have~$\bm{f}_{\mathrm{C}_{xy}}=\bm{0}$, where the subscript~$(\cdot)_{xy}$
denotes the tangential component. Substituting~\eqref{eq:GI_in_terms_of_ee}
into~\eqref{eq:obj_dynamics} with~$\bm{f}_{\mathrm{C}_{xy}}=\bm{0}$ and
dividing out~$m$ gives us $[\dot{\bm{v}}_e - \bm{R}_e\bm{g} +
(\dot{\bm{\omega}}^\times_e+\bm{\omega}^{\times}_e\bm{\omega}^{\times}_e)\bm{c}]_{xy}
= \bm{0}$. So far this is fine: we can plan trajectories that always satisfy
this equation. However, if we have two objects~$\mathcal{O}^{a}$
and~$\mathcal{O}^{b}$ with CoMs~$\bm{c}^{a}$ and~$\bm{c}^{b}$, respectively
(e.g., the left arrangement in Fig.~\ref{fig:support_planes}), then the EE
trajectory needs to satisfy \emph{both}
\begin{align}
  [\dot{\bm{v}}_e - \bm{R}_e\bm{g} + (\dot{\bm{\omega}}^\times_e+\bm{\omega}^{\times}_e\bm{\omega}^{\times}_e)\bm{c}^a]_{xy} = \bm{0},\label{eq:zero_fric_Oa} \\
  [\dot{\bm{v}}_e - \bm{R}_e\bm{g} + (\dot{\bm{\omega}}^\times_e+\bm{\omega}^{\times}_e\bm{\omega}^{\times}_e)\bm{c}^b]_{xy} = \bm{0},\label{eq:zero_fric_Ob}
\end{align}
at all times, where the only difference between~\eqref{eq:zero_fric_Oa}
and~\eqref{eq:zero_fric_Ob} is the CoM~$\bm{c}$. In general,
we cannot find an EE trajectory with non-zero accelerations that always satisfies both
equations. However, if there is \emph{some} friction force, the
right-hand sides of~\eqref{eq:zero_fric_Oa} and~\eqref{eq:zero_fric_Ob} are no
longer restricted to be identically zero and also need not be equal to each other.
Thus we choose to soften the object dynamics constraints (see the next
section), which allows tangential contact force to be used when needed, but
with a penalty. This approach still requires tuning: instead of
tuning friction coefficients, we now must tune the penalty weights.
The benefit is that we obtain computational savings when each force can be
represented by a non-negative scalar.

\section{Constrained Model Predictive Controller}\label{sec:mpc}

We now formulate a model predictive controller to solve the waiter's problem.
The controller optimizes trajectories~$\bm{x}(\tau)$, $\bm{u}(\tau)$,
and~$\bm{\xi}(\tau)$ over a time horizon~$\tau\in[t,t+T]$ by solving a nonlinear
optimization problem at each control timestep~$t$. Suppressing the time
dependencies, the problem is
\begin{equation}\label{eq:nonlinear_mpc_opt_prob}
  \begin{aligned}
    \argmin_{\bm{x},\bm{u},\bm{\xi}} &\quad \frac{1}{2}\int_{\tau=t}^{t+T} L(\bm{x},\bm{u},\bm{\xi})\ d\tau \\
    \text{subject to} &\quad \dot{\bm{x}} = \bm{a}(\bm{x})+\bm{B}(\bm{x})\bm{u} & \text{(system model)} \\
                      &\quad (\bm{e}(\bm{x}),\bm{\xi})\in\mathcal{B} & \text{(balancing)} \\
                      &\quad \bm{0} \leq \bm{d}(\bm{x}) & \text{(collision)} \\
                      &\quad \ubar{\bm{x}} \leq \bm{x} \leq \bar{\bm{x}} &\text{(state limits)} \\
                      &\quad \ubar{\bm{u}} \leq \bm{u} \leq \bar{\bm{u}} &\text{(input limits)}
  \end{aligned}
\end{equation}
where the stage cost is
\begin{equation*}
  L(\bm{x},\bm{u},\bm{\xi}) = \|\Delta\bm{r}(\bm{x})\|^2_{\bm{W}_r} + \|\bm{x}\|^2_{\bm{W}_x} + \|\bm{u}\|^2_{\bm{W}_u} + \|\bm{\xi}\|^2_{\bm{W}_f},
\end{equation*}
with~$\|\cdot\|^2_{\bm{W}}=(\cdot)^T\bm{W}(\cdot)$ for weight matrix~$\bm{W}$.
The EE position error
is~$\Delta\bm{r}(\bm{x})=\bm{r}_{d}-\bm{r}_{e}(\bm{x})$.
We focus on the case where the desired
position~$\bm{r}_{d}$ is constant, to assess the ability of our controller
to rapidly move to a new position without a priori trajectory information.
The matrices~$\bm{W}_r$ and~$\bm{W}_x$ are positive semidefinite;~$\bm{W}_u$
and~$\bm{W}_f$ are positive definite. Notice that we do not include a desired
orientation: we allow the balancing constraints to handle orientation as
needed. If~$\mu_i=0$, then only a scalar~$f_i$ is included as a decision
variable for each contact force (contained in~$\bm{\xi}$)
and~\eqref{eq:friction_cone} is replaced by the constraint~$f_i\geq 0$. The
vector~$\bm{d}(\bm{x})$ contains the distances between all pairs of collision
spheres representing obstacles and the robot body, which must be non-negative
to avoid collisions. When \emph{dynamic} obstacles are used, then we also
augment the state~$\bm{x}$ to predict their motion
(see~Sec.~\ref{sec:exp:dynamic}). We assume that the system can always reach a
feasible state that achieves the desired EE position. We discretize the
prediction horizon of~\eqref{eq:nonlinear_mpc_opt_prob} with a fixed
timestep~$\Delta t$ and solve it online using sequential quadratic programming
(SQP) via the open-source framework OCS2~\cite{ocs2}, with required Jacobians
computed using automatic differentiation. We assume that~$T$ can be chosen
sufficiently long to obtain stability. We use the Gauss-Newton approximation
for the Hessian of the cost and we soften the constraints with~$L_2$
penalties~\cite{frison2020hpipm}. The optimal state trajectory produced
by~\eqref{eq:nonlinear_mpc_opt_prob} is tracked by a low-level joint controller
at the robot's control frequency. More details can be found in the appendix.

\section{Simulation Experiments}\label{sec:sim}

We begin with simulations to gain insight into the performance of our
controller in an idealized environment. We use a simulated version of our
experimental platform, a $9$-DOF mobile manipulator consisting of a Ridgeback
mobile base and UR10 arm, depicted in Fig.~\ref{fig:experiment_setup}.
In all experiments (simulated and real) we use~$\Delta t=\SI{0.1}{s}$,
$T=\SI{2}{s}$, and weights
\begin{align*}
  \bm{W}_r &= \bm{I}_3, & \bm{W}_x &= \diag(0\bm{I}_9,0.1\bm{I}_9,0.01\bm{I}_9), \\
  \bm{W}_u &= 0.001\bm{I}_9, & \bm{W}_f &= 0.001\bm{I}_{\mathrm{dim}(\bm{\xi})},
\end{align*}
where~$\bm{I}_n$ is the~$n\times n$ identity matrix. We use a single SQP
iteration per control policy update.

\subsection{Balancing Constraint Comparison}

\begin{figure}[t]
  \centering
    \includegraphics[width=0.45\columnwidth]{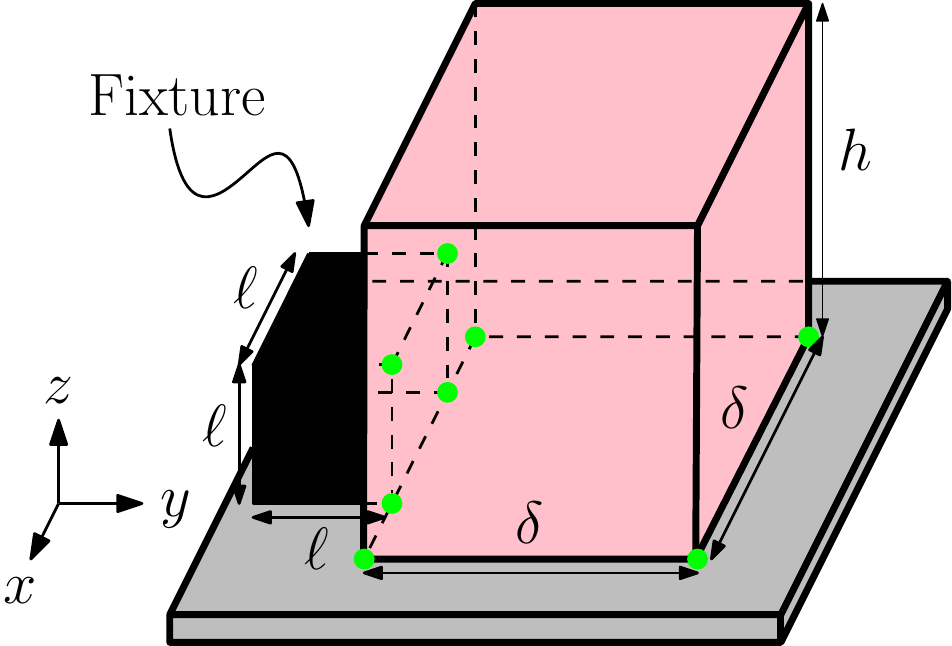}
    \caption{An arrangement consisting of a red box balanced on a tray along
    with a black \emph{fixture}, which is rigidly attached to the tray. The
    fixture adds contact points (shown in green) up the side of the box, which
    our controller can exploit to accelerate faster.}
  \label{fig:sim_example}
\end{figure}

We first consider the example shown in Fig.~\ref{fig:sim_example},
consisting of a box balanced on a tray and in contact with a \emph{fixture},
which is rigidly attached to the tray. We perform experiments with and without
the fixture, which is a cube of side length~$\ell=\SI{5}{cm}$. The box has
mass~$m=\SI{0.5}{kg}$, height~$h=\SI{20}{cm}$, and a square base with side
length~$\delta=\SI{6}{cm}$. The CoM is located at the centroid, the mass
distribution is uniform, and~$\mu_i=0.2$ for all~$i\in\mathcal{I}$. The task
is to move the EE to a desired goal point~$\bm{r}_d=[-2,1,0]^T$ (all desired
positions are given in meters relative to the initial EE position) without
dropping the box. We compare the trajectories that result from imposing four
different sets of balancing constraints:
\begin{itemize}
  \item \textbf{None:} No constraints.
  \item \textbf{Upward:} A constraint to keep the tray oriented upward.
  \item \textbf{Full:} The full set of balancing
    constraints~$(\bm{e},\bm{\xi})\in\mathcal{B}$ with each~$\mu_i$ set to
    $90\%$ of the true value.\footnote{We only use~$90\%$ of the true (measured
    or simulated) value to provide some robustness to small constraint
    violations arising from discretization errors and other numerical
    disturbances. We subtract a small margin from the support area for the
    same reason. This is more important in the hardware experiments, where
    there are more sources of noise and disturbances.}
  \item \textbf{Robust:} The full set of constraints~$(\bm{e},\bm{\xi})\in\mathcal{B}$
    with $\{\mu_i\}_{i\in\mathcal{I}}$ computed using~\eqref{eq:minimum_mu}.
    Unless otherwise stated, the solution is~$\mu_i=0$ for all
    $i\in\mathcal{I}$.
\end{itemize}
In ideal conditions, the Full and Robust constraints should both keep the
objects balanced, but the Full constraints provide less of a safety margin. The
Upward approach would work if the motion were quasistatic (i.e., with negligible
accelerations), but that would not be fast or reactive.

\begin{figure}[t]
  \centering
    \includegraphics[width=\columnwidth]{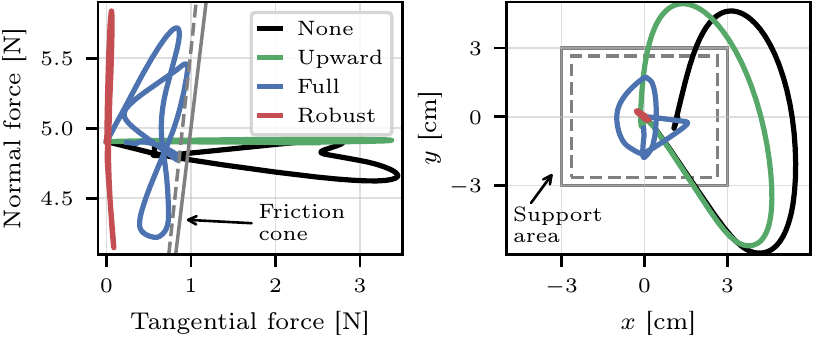}
    \caption{\emph{Left:} Force applied to the simulated box during motion.
    \emph{Right:} Corresponding ZMP trajectories.  With no constraints
    (None) or the Upward constraint, the force
    leaves the friction cone and the ZMP leaves the support area (safety
    margins in dashed lines), so the box slides and tips over (and is dropped).
    The Full constraints touch but do not pass the boundaries;
    the Robust constraints stay far from the boundaries in both cases.}
  \label{fig:fric_cone_support_area}
\end{figure}

\begin{figure}[t]
  \centering
    \includegraphics[width=\columnwidth]{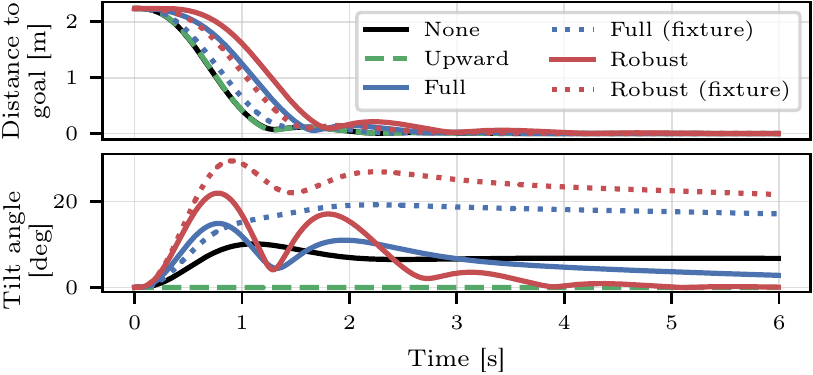}
    \caption{\emph{Top:} Distance of EE to goal location. \emph{Bottom:} Tilt
    angle with respect to the upward-pointing (i.e. gravity-aligned)
    orientation. The Full and Robust constraints limit acceleration to keep the
    box balanced; the Robust approach also uses higher tilt angles. The None
    and Upward approaches accelerate faster---and drop the box. When the
    fixture is added, the Full and Robust constraints can exploit it to achieve
    convergence speeds more similar to the None and Upward cases. Notice that,
    except for the Upward constraint, there is no need for the tilt angle to be
    near zero.}
  \label{fig:simulation_trajectory}
\end{figure}

In Fig.~\ref{fig:fric_cone_support_area}, the force acting on the
object and the zero-moment point (ZMP) are shown relative to the friction cone and support area,
respectively. The ZMP is the point about which horizontal moments
are zero; if it is outside of the support area, then the object tips.
Unsurprisingly, the None and Upward approaches significantly violate both the
friction cone and ZMP constraints, resulting in the box being dropped. The Full
approach produces motion at the boundary of the constraints but does not
violate them, while the Robust approach stays away from the boundaries. In
Fig.~\ref{fig:simulation_trajectory}, we see that the robustness of the Robust
approach comes at the cost of slower convergence and higher tilt angles
compared to the Full approach. When the fixture is added, the Full and Robust
approaches can both exploit it to converge nearly as fast as when no
constraints are used at all. Notice that the tilt angle for the None
and Full approaches need not converge to zero: the None approach does not
consider the EE orientation at all, while the Full approach may converge to any
orientation that satisfies the balancing constraints.

\subsection{Non-Parallel Support Planes}\label{sec:sim:nonparallel}

\begin{figure}[t]
  \centering
    \includegraphics[width=0.32\columnwidth,trim={0 0 0 0},clip]{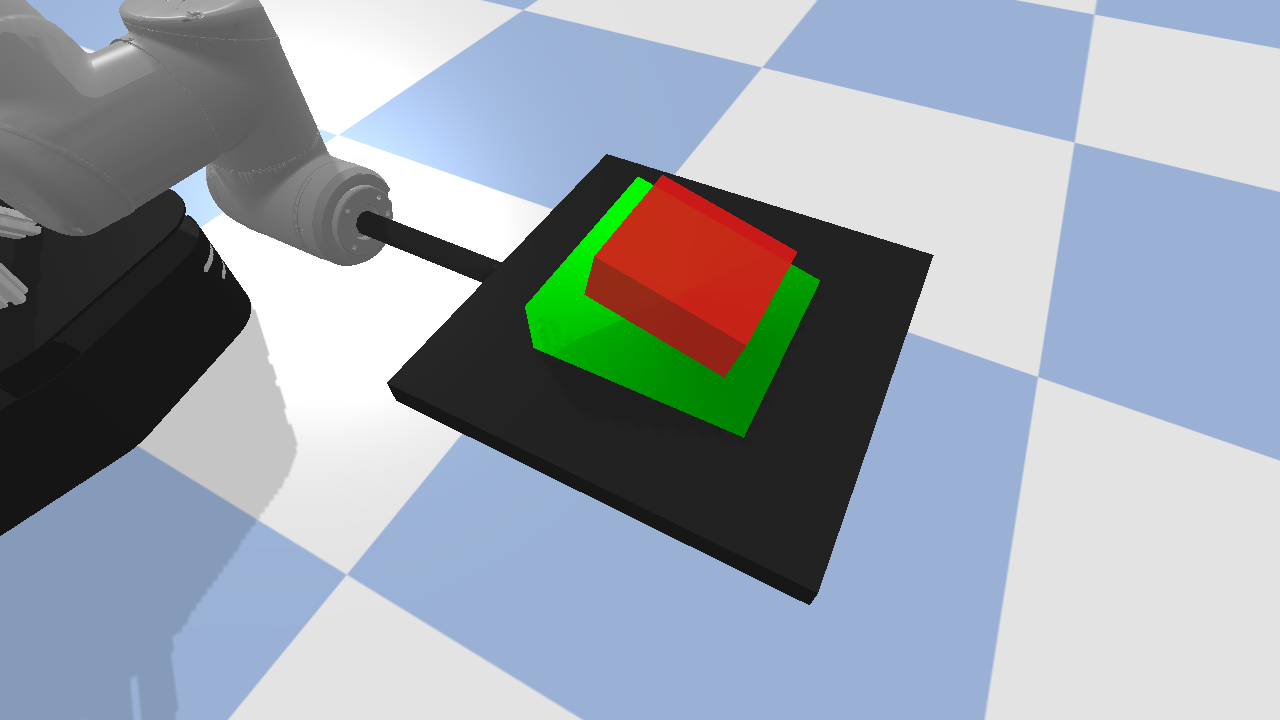}
    \hfill
    \includegraphics[width=0.32\columnwidth,trim={0 0 0 0},clip]{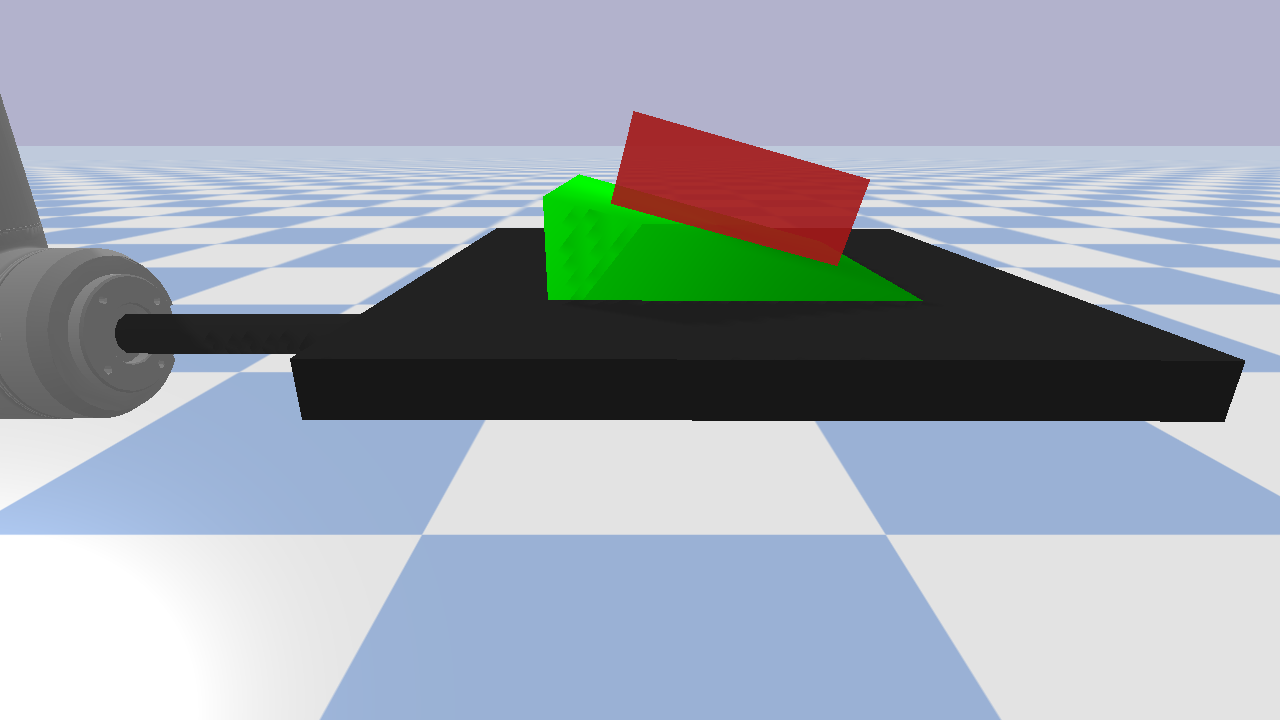}
    \hfill
    \includegraphics[width=0.32\columnwidth,trim={0 0 0 0},clip]{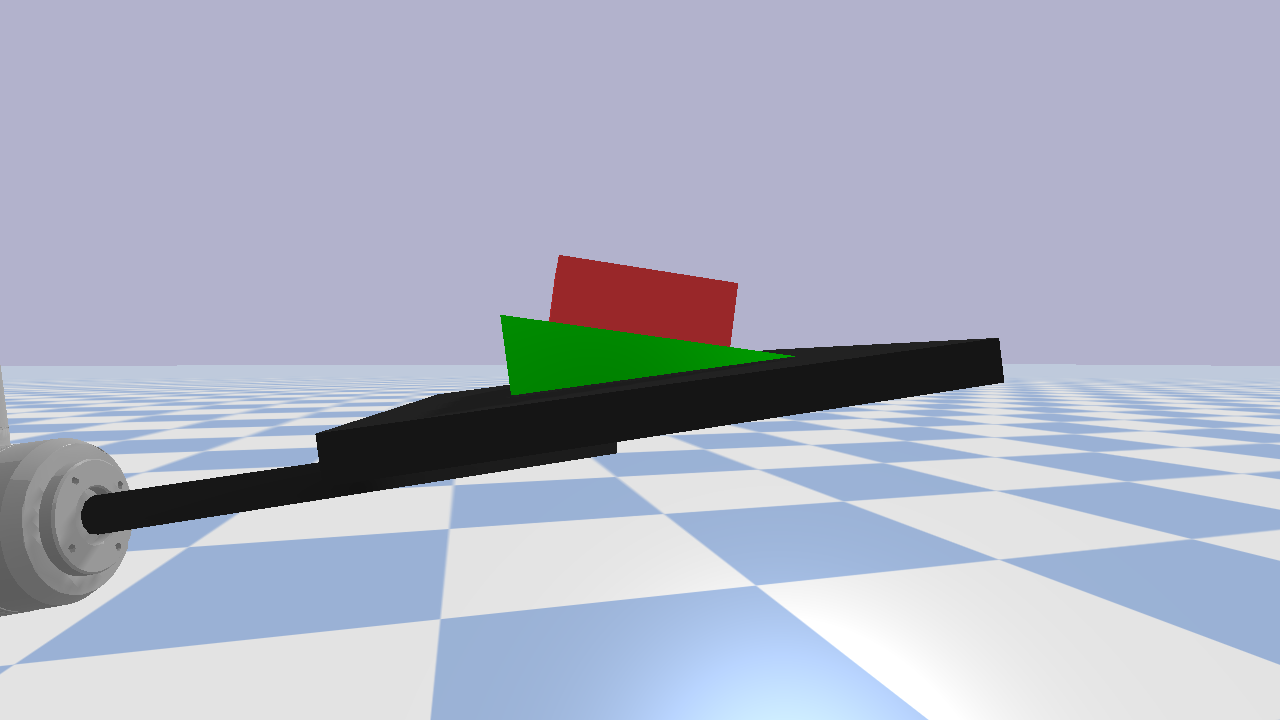}
    \caption{\emph{Left:} Initial position of wedge (green) and box (red)
    arrangement. \emph{Middle:} Initial side view. The box is
    tilted~\SI{15}{\degree} relative to the ground due to the slope of the wedge.
    \emph{Right:} The position at~$t=\SI{6}{s}$. The controller has
    oriented the tray so that both the wedge and box are tilted
    \SI{7.5}{\degree} relative to the ground, such that each requires as small
    a~$\mu$ as possible.}
  \label{fig:wedge}
\end{figure}

Next we show an example when the solution to~\eqref{eq:minimum_mu} is not
simply~$\mu_i=0$ for all~$i\in\mathcal{I}$. The setup consists of a wedge
supporting a box at an incline of~$\phi=\SI{15}{\degree}$, similar to the right side of
Fig.~\ref{fig:support_planes}. For simplicity we assume that~$\mu$ is constant
between each pair of objects, so we need only solve~\eqref{eq:minimum_mu} for
the friction coefficient between the tray and wedge~$\mu_{tw}$ and between the
wedge and box~$\mu_{wb}$. We obtain~$\mu_{tw}=\mu_{wb}=0.132$, which
corresponds to a tilt angle of~$\theta=\arctan(0.132)\approx\SI{7.5}{\degree}$
relative to the ground for each object, meaning the wedge and box can be oriented so as to split the
angle~$\phi$ between them. Using~$\mu_{tw}=\mu_{wb}=0.132$ for the controller
and~$\mu_{tw}=\mu_{wb}=0.2$ for the simulator, we run the simulation with the
same goal~$\bm{r}_d=[-2,1,0]^T$. The initial and final configurations are shown
in Fig.~\ref{fig:wedge}. Note that we need not start in a configuration which
the controller thinks is feasible (since the constraints are soft), but the
controller will steer toward one over the course of the trajectory. If we were
to dispense with~\eqref{eq:minimum_mu} and simply try to enforce~$\mu_i=0$ for
all~$i\in\mathcal{I}$, the controller fails to converge because no feasible
stationary solution exists.

\section{Hardware Experiments}\label{sec:exp}

\begin{figure}[t]
  \centering
    \includegraphics[height=0.95in,trim={0 0 0 0},clip]{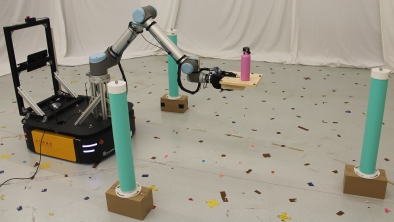}
    \hfill
    \includegraphics[height=0.95in,trim={0 0 0 0},clip]{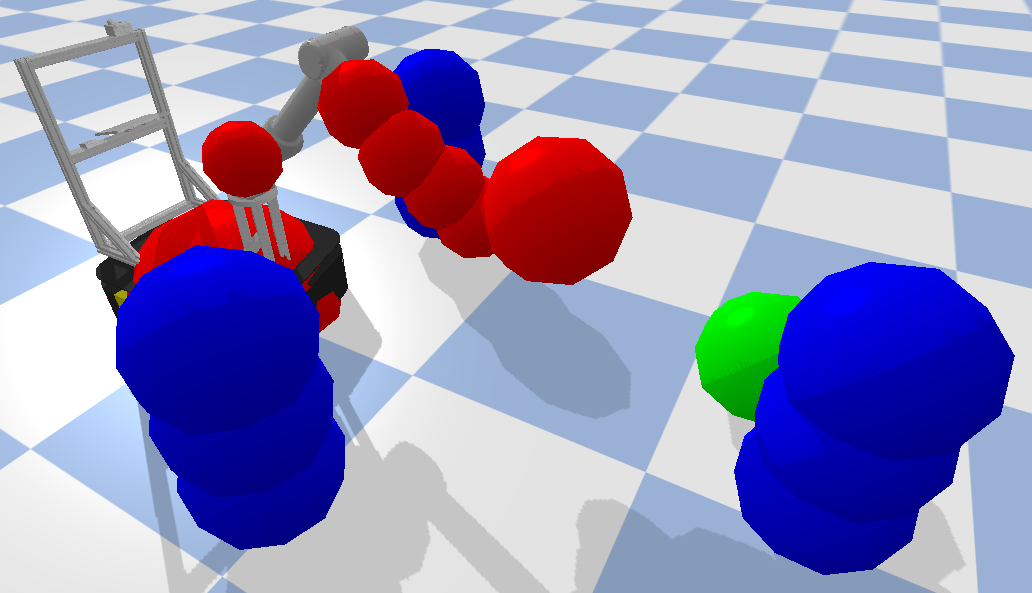}\\[1ex]
    \includegraphics[height=0.72in,trim={0 0 0 0},clip]{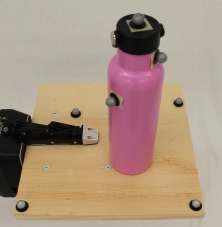}
    \includegraphics[height=0.72in,trim={0 0 0 0},clip]{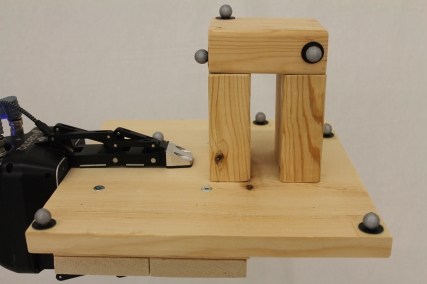}
    \includegraphics[height=0.72in,trim={0 0 0 0},clip]{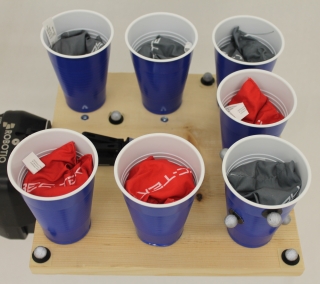}
    \caption{\emph{Top left:} Real experimental setup. Robot is shown holding
    the Bottle object. Obstacle locations marked with pylons. \emph{Top right:}
    Corresponding simulated experimental setup with collision spheres on the
    robot in red and on the obstacles in blue. \emph{Bottom row:} Bottle,
    Arch, and Cups object arrangements used for experiments. The arch is an
    example of non-coplanar contact (the three blocks composing the arch are
    not attached together). The bottle is filled with sugar and the cups each
    contain bean bags instead of liquid to avoid spills in the lab.}
  \label{fig:experiment_setup}
  \vspace{10pt}
\end{figure}

In simulation we gained insight into the behaviour of the controller without
the influence of real-world effects like sensor noise or EE vibrations. We now
perform experiments on our real mobile manipulator to assess our approach in
more realistic scenarios. Position feedback is provided for the arm by joint
encoders and for the base by a Vicon motion capture system, which is used in a
Kalman filter to estimate the full robot state. We also use motion capture to
track the position of the balanced objects, which is only used
for error reporting. The controller parameters and weights are the same as in
the previous section. The controller is run on a standard laptop with eight
Intel Xeon CPUs at~\SI{3}{GHz} and~\SI{16}{GB} of RAM. The robot and balanced
objects are shown in Fig.~\ref{fig:experiment_setup}; the corresponding object
parameters are given in Table~\ref{tab:object_parameters}. A video of the
experiments can be found at \texttt{\small \videourl}.

\begin{table}[t]
  \caption{Approximate parameters for balanced objects shown in
  Fig.~\ref{fig:experiment_setup}. CoM and inertia are estimated from mass and
  geometry.}
  \centering
  \begin{tabular}{c c c c l}
    \toprule
    Arrangement & \makecell{\# of\\objects} & \makecell{\# of\\contacts} & \makecell{Mass per\\object [g]} & \makecell{Friction\\coefficients} \\
    \midrule
    Bottle & 1 & 4 & 827 & \emph{tray-bottle:} 0.26 \\
    \midrule
    Arch   & 3 & 16 & 180 & \makecell[l]{\emph{tray-block:} 0.30\\\emph{block-block:} 0.42} \\
    \midrule
    Cups   & 7 & 28 & 200 & \emph{tray-cup:} 0.28 \\
    \bottomrule
  \end{tabular}
  \label{tab:object_parameters}
\end{table}

\subsection{Static Environments}

\begin{figure}[t]
  \centering
    \includegraphics[width=\columnwidth]{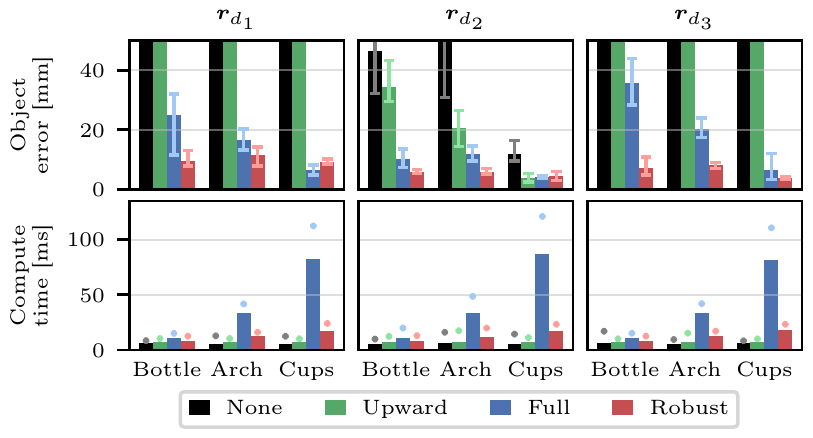}
    \caption{Object error (top row) and policy compute time (bottom row) for
      different combinations of objects, goal positions, and constraints in
      free space. The object error is the maximum distance the object moves from
      its initial position relative to the tray.  In arrangements with multiple
      objects, only a single one is tracked. The bar shows the average of
      three runs; the error bars show the minimum and maximum values. One or more
      objects were completely dropped in all cases where the minimum error is beyond
      the axis limits. The compute time is the average time required
      to compute an updated MPC policy (i.e., one iteration
      of~\eqref{eq:nonlinear_mpc_opt_prob}). The bar shows the average across
      the three runs (up to the first~\SI{6}{s} of the trajectory);
      the dot shows the maximum value from any of the runs.}
  \label{fig:freespace}
\end{figure}

\begin{figure}[t]
  \centering
    \includegraphics[width=\columnwidth]{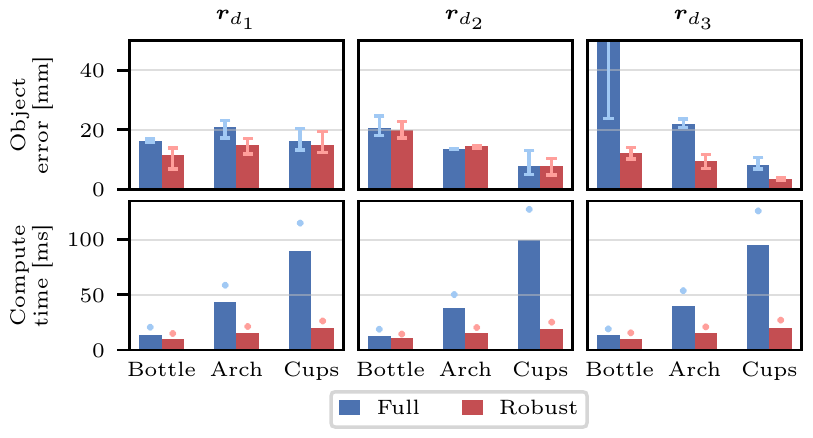}
    \caption{The same results as shown in Fig.~\ref{fig:freespace} but in an
      environment with static obstacles and only showing the Full and Robust
      approaches. Compared to Fig.~\ref{fig:freespace}, the errors are similar
      while the compute times are slightly higher.}
  \label{fig:static_obstacles}
\end{figure}

We perform a large set of experiments with different combinations of objects
and desired EE positions, each using the None, Upward, Full, and Robust
constraint methods described above. The desired positions
are~$\bm{r}_{d_1}=[-2,1,0]^T$, $\bm{r}_{d_2}=[2,0,-0.25]^T$,
and~$\bm{r}_{d_3}=[0,2,0.25]^T$. The object error and controller compute time
in an obstacle-free environment are shown in Fig.~\ref{fig:freespace}; results
for an environment with static obstacles are shown in
Fig.~\ref{fig:static_obstacles}. We model obstacles as collections of spheres;
spheres also surround parts of the robot body for collision checking (see top
right of Fig.~\ref{fig:experiment_setup}). As expected, the None and Upward
approaches almost always fail---the notable exception is for
goal~$\bm{r}_{d_2}$, which requires more base motion and is thus slower than
the other trajectories. The Robust constraints typically produce the lowest
object error or are close to it. In general we expect the Robust constraints to
have the lowest error, given that they reduce the tangential contact forces and
can thus resist unmodelled force disturbances. However, we noticed that the
larger tilt angles required by the Robust constraints can occasionally result
in some sliding of the objects.

\begin{figure}[t]
  \centering
    \includegraphics[width=\columnwidth]{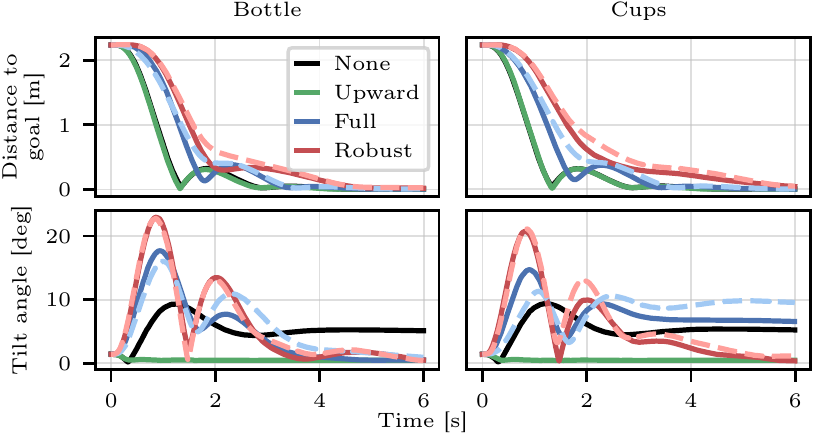}
    \caption{Samples of trajectories to goal~$\bm{r}_{d_1}$ for the Bottle
      and Cup arrangements with different constraints. Free space results are
      solid lines; results with static obstacles (shown only for Full
      and Robust) are dashed. The addition of static obstacles modestly
      changes the shape of the trajectories. The Full and Robust trajectories
      differ between the two object arrangements; in particular, notice that
      Full constraints converge to a much smaller tilt angle with the Bottle
      compared to the Cups. The Bottle's higher CoM makes it easier to tip, so it
      requires a smaller tilt angle when stationary.}
  \label{fig:static_env_trajectories}
\end{figure}

Computationally the Robust constraints scale \emph{much} better with the number
of contacts than the Full constraints, since they require less decision
variables and use simpler constraints. The Full constraints also require
reasonably accurate friction coefficient estimates; the effectiveness of the
Robust constraints show that we need not fear frictional uncertainty and (when
statically feasible) can set~$\mu_i=0$ for all~$i\in\mathcal{I}$ to reduce
compute time. The static obstacle results in
Fig.~\ref{fig:static_obstacles} are similar to those for free space except for
a modest increase in compute time. Sample trajectories are shown in
Fig.~\ref{fig:static_env_trajectories}.

\subsection{Dynamic Environments}\label{sec:exp:dynamic}

We now consider environments that change over time due to dynamic obstacles.
Dynamic obstacles are modelled as spheres with known radii, but the controller
does not know their trajectories a priori.

\subsubsection{An Unexpected Obstacle}

\begin{figure}[t]
  \centering
    \includegraphics[width=\columnwidth]{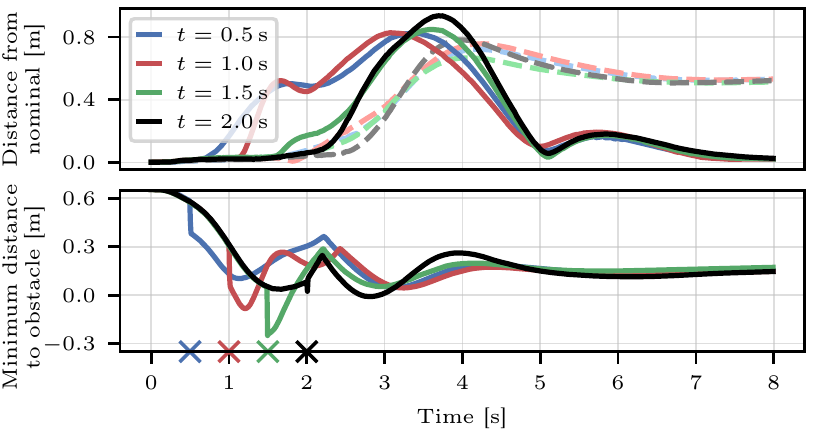}
    \caption{\emph{Top:} Distance of EE (solid) and base (dashed) positions
    from a nominal trajectory with a virtual obstacle suddenly appearing at different
    times~$t$ (also marked with crosses on $x$-axis). The nominal trajectory
    has no dynamic obstacle. \emph{Bottom:} Minimum distance between any
    collision sphere on the robot and the dynamic obstacle. Notice that in some
    cases the appearance of the obstacle actually violates the collision
    constraints, which could also happen with a physical obstacle if the
    collision sphere was conservatively large. Regardless, the Bottle was
    never dropped.}
    \label{fig:sudden_obstacle_trajectories}
\end{figure}

Here we test the controller's ability to react to unexpected events. We make
the controller aware of a new obstacle at varying times~$t$, and the policy
must be quickly updated to avoid a collision. The setup is simple: we use the
static obstacle environment and goal~$\bm{r}_{d_2}$ with the Bottle
arrangement and a new ``virtual'' obstacle (the obstacle does not physically
exist, but the controller thinks it is present). At time~$t$ the new
obstacle instantly appears in front of the robot (represented by the green sphere in
Fig.~\ref{fig:experiment_setup})---imagine a restaurant customer suddenly
backing out their chair. The results for different~$t$ are shown in
Fig.~\ref{fig:sudden_obstacle_trajectories}. The appearance of the obstacle
causes significant changes in the trajectory of both the EE and base, but the
object is continually balanced despite the sudden change, even when the
collision constraint is violated by the obstacle's appearance. The
maximum object error and policy compute time were~\SI{18}{mm} and~\SI{23}{ms},
respectively, across three runs of each of the four obstacle appearance
times~$t$. The trajectory with~$t=\SI{1}{s}$ achieved the highest EE velocity
and acceleration of all our experiments, at~\SI{2.0}{m/s}
and~\SI{7.9}{m/s\squared}, respectively.

\subsubsection{Projectile Avoidance}

\begin{figure}[t]
  \centering
  \footnotesize
  \stackunder[5pt]{\includegraphics[height=0.67in]{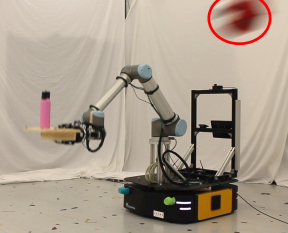}}{$t$}%
  \stackunder[5pt]{\includegraphics[height=0.67in]{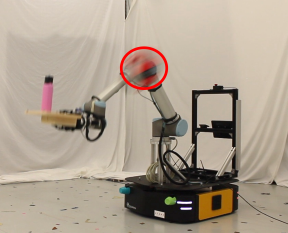}}{$t+\Delta$}%
  \stackunder[5pt]{\includegraphics[height=0.67in]{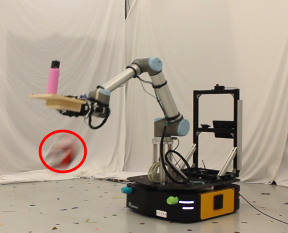}}{$t+2\Delta$}%
  \stackunder[5pt]{\includegraphics[height=0.67in]{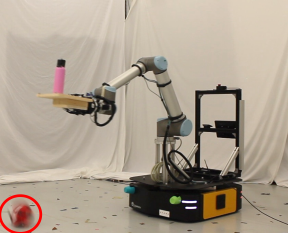}}{$t+3\Delta$}%
  \caption{Example of the robot dodging the volleyball (circled red) while
  balancing the bottle, with frames spaced by~$\Delta=\SI{0.15}{s}$. Once the
  ball has passed, the EE moves back to the initial position.}
  \label{fig:projectile_stop_motion}
\end{figure}

\begin{figure}[t]
  \centering
    \includegraphics[width=\columnwidth]{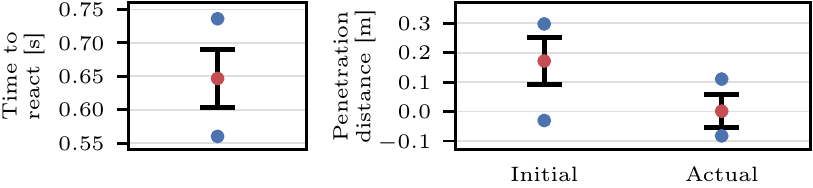}
    \caption{The projectile avoidance results over 20
      trials. In each plot the red dot is the mean, the error bars represent
      the standard deviation, and the blue dots are the minimum and
      maximum values. \emph{Left:} The time at which collision would first
      occur if the robot did not move.
      In all cases the controller has less than~\SI{0.75}{s} to react.
      \emph{Right:} Maximum penetration distance between the (virtual)
      collision spheres around the ball and EE. The ``Initial'' values
      represent the maximum penetration distances that would have occurred if
      the robot had not moved. The ``Actual'' values are what really happened
      given that the robot did move.}
  \label{fig:projectile}
\end{figure}

Finally, we consider a ball with position~$\bm{r}_{b}$ and
state~$\bm{b}=[\bm{r}_{b}^T,\dot{\bm{r}}_{b}^T]^T$ modelled as
a simple projectile with~$\ddot{\bm{r}}_{b} = \bm{g}$.
We neglect drag and other possible nonlinear terms, because \emph{avoiding} an
object requires a less accurate model than when
catching~\cite{dong2020catch} or
batting it~\cite{zhang2018real}. The ball is thrown toward the EE, and the
robot must move to avoid the objects being hit while also keeping them
balanced. For these experiments we use the Bottle arrangement and
the Robust constraint method. The controller is provided with feedback
of~$\bm{b}$ once the ball exceeds the height of~\SI{1}{m}; the state is
estimated using the motion capture system. The state~$\bm{b}$ and the
projectile dynamics are added to~\eqref{eq:nonlinear_mpc_opt_prob} to predict
the ball's motion. We found it most effective to use a form of continuous
collision checking in which the controller tries to avoid a tube
around the entire future trajectory of the ball. Once the ball has passed the
EE, the constraint is removed.

The results for 20 throws are shown in Fig.~\ref{fig:projectile} and images from one throw are shown in Fig.~\ref{fig:projectile_stop_motion}. Throws are
split evenly between two directions: toward the front of the EE and toward its
side. In all cases, the controller has less than~\SI{0.75}{s} to react and
avoid the ball. Out of the 20 trials, there is one in which the ball would not
have penetrated the collision sphere even if the robot did not move, and
another where the bottle was actually dropped. This failure was not due
to a collision, but because the bottle tipped over due to the aggressive motion
used to avoid the ball. Also notice that the controller does not always
completely pull the robot out of collision: there is a trade-off between
balancing the object and avoiding collision.
However, since
the collision spheres are conservatively large, we did not experience any
failures due to collisions. In these experiments the controller only tries to
avoid collisions between the ball and EE; collisions with the rest of the
robot's body are not avoided.
The maximum object error and policy compute time
were~\SI{32}{mm} (ignoring the single failure) and~\SI{20}{ms}, respectively,
across the 20 trials.

\section{Conclusion}\label{sec:conclusion}

We presented an MPC-based approach for balancing objects with a
velocity-controlled mobile manipulator and demonstrated its performance in
simulated and real experiments in a variety of static and dynamic scenarios. In
particular, our method is able to react quickly to moving obstacles. We also
proposed using minimal values of~$\mu$ to add robustness to frictional
uncertainty and other force disturbances, and demonstrated that this approach
is effective and computationally efficient in real-world experiments.
Future work will explore the effect of uncertainty in the objects' inertial
parameters and the use of object state feedback in the controller.

\appendix

This appendix provides additional implementation details to complement the main
body of the manuscript.

\subsection{Robot Kinematic Model}

In the main body of the manuscript we leave the robot kinematic model general
to accommodate different systems. The actual kinematic model for the robot used
in our experiments is
\begin{equation}\label{eq:robot_model_ct}
  \dot{\bm{x}} = \bm{A}\bm{x} + \bm{B}\bm{u},
\end{equation}
where
\begin{align*}
  \bm{A} &= \begin{bmatrix}
    \bm{0}_9 & \bm{I}_9 & \bm{0}_9 \\
    \bm{0}_9 & \bm{0}_9 & \bm{I}_9 \\
    \bm{0}_9 & \bm{0}_9 & \bm{0}_9 \\
  \end{bmatrix}, &
  \bm{B} &= \begin{bmatrix}
    \bm{0}_9 \\ \bm{0}_9 \\ \bm{I}_9
  \end{bmatrix},
\end{align*}
with~$\bm{0}_n$ denoting a~$n\times n$ matrix of zeros. The fact that our
mobile base is omnidirectional gives us a linear model, but the nonlinear
equations of motion arising from a nonholonomic base, for example, can also be
handled, since the MPC problem~(9) is already nonlinear.

\subsection{Minimum Statically-Feasible Friction Coefficients}

Here we provide more details on the structure of the optimization problem~(6),
used to obtain the minimum statically-feasible friction coefficients. For
simplicity we assume that a single object is balanced. We use
roll-pitch-yaw Euler angles~$\bm{\theta}$ to parameterize rotation. In this
case~(6) has the form
\begin{subequations}\label{eq:minimum_mu_ex}
  \begin{align}
    \argmin_{\bm{\theta},\bm{\xi},\{\mu_i\}_{i\in\mathcal{I}}} &\quad \frac{1}{2}\sum_{i\in\mathcal{I}}\alpha_i\mu_i^2 \\
    \text{subject to} &\quad \mu_i \geq 0, \quad i\in\mathcal{I} \\
                        &\quad \bm{F}_i\bm{f}_i\geq\bm{0}, \quad i\in\mathcal{I}\label{eq:minimum_mu_ex:friccone} \\
                        &\quad \sum_{i\in\mathcal{I}}\begin{bmatrix} \bm{f}_i \\ \bm{r}_i\times\bm{f}_i \end{bmatrix} = \begin{bmatrix} -m\bm{R}_e(\bm{\theta})\bm{g} \\ \bm{0} \end{bmatrix},\label{eq:minimum_mu_ex:ne}
  \end{align}
\end{subequations}
where~$\bm{R}_e(\bm{\theta})$ is the mapping from~$\bm{\theta}$ to the
corresponding rotation matrix and
\begin{equation*}
  \bm{F}_i = \begin{bmatrix}
    1 & 0 & 0 \\
    \mu_i & -1 & -1 \\
    \mu_i & -1 & 1 \\
    \mu_i & 1 & -1 \\
    \mu_i & 1 & 1
    \end{bmatrix}\begin{bmatrix} \hat{\bm{n}}_i^T \\ \bm{S}_i^T \end{bmatrix}
\end{equation*}
with~$\bm{S}_i$ an orthonormal basis for the tangential component
of~$\bm{f}_i$, such that~\eqref{eq:minimum_mu_ex:friccone} is equivalent to the
friction cone constraint~(3). This is the same friction model that appears in,
e.g.,~[21]. For multiple objects, each would need a set of Newton-Euler equality
constraints~\eqref{eq:minimum_mu_ex:ne}. We would also need to include the
constraint~(5) for any contact points between two of the objects.

The problem~(6) is always non-convex due to the product of decision
variables~$\mu_i$ and~$\bm{f}_i$ in the friction cone constraint and the
nonlinear mapping~$\bm{R}_e(\bm{\theta})$, but we did not have a problem
solving it with the SLSQP solver~\cite{kraft1988a} from \texttt{scipy}.

\subsection{Soft Constraints}

Here we provide details the soft constraints used in~(9). We soften all of the
constraints in~(9) except for the system model
constraints~$\dot{\bm{x}}=\bm{a}(\bm{x})+\bm{B}(\bm{x})\bm{u}$.

Consider a general inequality constraint~$g(\bm{x},\bm{u})\leq0$ (equality
constraints are just treated as two-sided inequalities with equal lower and
upper limits). We \emph{soften} the constraint by introducing a slack
variable~$s\geq0$ as another decision variable in the optimization problem and
relaxing the inequality constraint to~$g(\bm{x},\bm{u})\leq s$. The optimizer
is encouraged to make~$s$ small (and thus reduce constraint violation) by
adding a term penalizing~$s$ to the objective function. In this work we use
an~$L_2$ penalty of the form~$w_ss^2$ for each slack variable, where~$w_s>0$ is
a tunable weight. We use~$w_s=100$ for all slacks except for the projectile
avoidance constraint, which uses~$w_s=4/d^2$, where~$d=\SI{0.35}{m}$ is the
specified minimum distance between the end effector and the predicted
projectile trajectory. We found that the relative weight between the slack
penalties for the balancing constraints and the projectile avoidance
constraints was the most difficult part of the controller to tune, hence the
different slack weight for the projectile avoidance constraint.


When the constraints are soft, the relative magnitudes of the constraint
violations must also be considered (which are weighed against each other in the
problem's objective function). In particular, we adjust the Newton-Euler
dynamics constraints~(1) for each object to
\begin{equation*}
  m^{-1}\left(\bm{w}_{\mathrm{C}} + \bm{w}_{\mathrm{GI}}/\sqrt{N}\right) = \bm{0}.
\end{equation*}
Dividing by the object's mass~$m$ ensures that balancing heavier objects is not
prioritized over lighter objects. Dividing the gravito-inertial wrench
by~$\sqrt{N}$ reduces the magnitude of the contact force variables in the
optimization problem as the number of contact points~$N$ increases. The idea is
that we do not want the penalties on~(1) to dominate the other objectives and
penalties in the problem~(9) just because more objects and contact points have
been added to the problem. We found this to be particularly useful with the
Cups arrangement (where~$N=28$).

\subsection{Model Predictive Controller Details}

Here we provide some additional details about the MPC problem~(9). The state
and input constraints used for the robot in~(9) are
\begin{alignat*}{4}
  \bar{\bm{q}} &= \begin{bmatrix} 10\bm{1}_3 \\ 2\pi\bm{1}_6 \end{bmatrix}, &\;
  \bar{\bm{v}} &= \begin{bmatrix} 1.1\bm{1}_2 \\ 2\bm{1}_3 \\ 3\bm{1}_4 \end{bmatrix}, &\;
  \dot{\bar{\bm{v}}} &= \begin{bmatrix} 2.5\bm{1}_2 \\ 1 \\ 10\bm{1}_6 \end{bmatrix}, &\;
  \bar{\bm{u}} &= \begin{bmatrix} 20\bm{1}_3 \\ 80\bm{1}_6 \end{bmatrix},
\end{alignat*}
where~$\bar{\bm{x}}=[\bar{\bm{q}}^T,\bar{\bm{v}}^T,\dot{\bar{\bm{v}}}^T]^T$,
$\ubar{\bm{x}}=-\bar{\bm{x}}$, $\ubar{\bm{u}}=-\bar{\bm{u}}$, and~$\bm{1}_n$
denotes an $n$-dimensional vector of ones.

As stated in the main manuscript, (9) is solved using sequential quadratic
programming. The quadratic program (QP) subproblems are solved using the QP
solver HPIPM~[24]. The automatic differentiation library CppAD~\cite{cppad} is
used to obtain the Jacobians required to construct the QP subproblems.

\subsection{Low-level Joint Controller}

The MPC problem~(9) typically cannot be solved at the same frequency that the
robot accepts commands, so we need a strategy to compute inputs between
solutions of~(9). Suppose we compute a new MPC policy using~(9) at time~$t$,
which is valid until time~$t+T$. Then at each control time~$\tau\in[t,t+T]$, we
can compute the jerk input~$\bm{u}(\tau)$ using an affine state feedback
controller of the form
\begin{equation}\label{eq:feedback_ctrl}
  \bm{u}(\tau) = \bm{K}(\tau)(\bm{x}^\star(\tau)-\bm{x}(\tau)) + \bm{k}(\tau),
\end{equation}
where~$\bm{x}^{\star}$ is the optimal state trajectory,~$\bm{K}$ is the
feedback gain matrix, and~$\bm{k}$ is the feedforward input, all obtained from
the most recent policy. In particular, at each control timestep~$t$, (9) is
discretized and linearized to form a quadratic program (QP), which is solved
using an interior point method (IPM)~[24]. The terms~$\bm{K}$ and~$\bm{k}$ are
obtained from the Riccati recursion used to solve the linear system arising
from the Karush-Kuhn-Tucker conditions in the final iteration of the IPM used
to solve the QP (see~[24] as well as~\cite{frison2014high}
and~\cite{rao1998application} for more details). The upshot
of~\eqref{eq:feedback_ctrl} is that we can cheaply generate inputs~$\bm{u}$
based on the most recent MPC solution, unless more time than the horizon~$T$
has elapsed since the solution, which never occurred during our experiments.

In simulation we do not run in real time, which allows us to recompute the
policy every~\SI{10}{ms} of simulation time, regardless of the actual required
compute time. We use~\eqref{eq:feedback_ctrl} to generate the input at every
step of the simulation, which has a timestep of~\SI{1}{ms}. In our hardware
experiments, the MPC policy~(9) is solved in a separate process. We limit
policy updates to at most once every~\SI{10}{ms} and we
use~\eqref{eq:feedback_ctrl} to generate commands at the robot's control
frequency of~\SI{125}{Hz}.

\subsection{State Estimation}

Two Kalman filters are used for state estimation in our hardware experiments.
One is used to estimate the state of the robot~$\bm{x}$ and the other is used
to estimate the state of the projectile~$\bm{b}$ during the dynamic obstacle
avoidance experiments in~Sec.~VIII-B.2. Both the robot and projectile models
are linear, allowing us to use the standard linear Kalman filter (see
e.g.~\cite{barfoot2017state}). We are provided with position measurements for
both systems: for the robot, the pose of the mobile base is provided by a Vicon
motion capture system, and the joint angles of the arm are provided by its
joint encoders. The position of the projectile is also obtained from the Vicon
system. In the following, we describe the discrete-time equations of motion and
the covariance matrices required for each Kalman filter.

\subsubsection{Robot Kalman Filter}

Discretizing~\eqref{eq:robot_model_ct} gives us the discrete-time model
\begin{equation*}
  \bm{x}^+ = \skew5\bar{\bm{A}}\bm{x} + \bar{\bm{B}}\bm{u},
\end{equation*}
where
\begin{align*}
  \skew5\bar{\bm{A}} &= \begin{bmatrix}
    \bm{I}_9 & \delta t\bm{I}_9 & (1/2)\delta t^2\bm{I}_9 \\
    \bm{0}_9 & \bm{I}_9 & \delta t\bm{I}_9 \\
    \bm{0}_9 & \bm{0}_9 & \bm{I}_9
  \end{bmatrix}, &
  \bar{\bm{B}} &= \begin{bmatrix}
    (1/6)\delta t^3\bm{I}_9 \\ (1/2)\delta t^2\bm{I}_9 \\ \delta t\bm{I}_9
  \end{bmatrix},
\end{align*}
are obtained from Taylor series expansions of~$\bm{x}$ and we have
used~$\bar{(\cdot)}$ to denote the discrete-time system matrices. The
sampling time is~$\delta t=\SI{8}{ms}$, which is the duration of each iteration
of the robot control loop. We measure generalized positions~$\bm{q}$, and so
our measurement model is~$\bm{q}=\bar{\bm{C}}\bm{x}$, where
\begin{equation*}
  \bar{\bm{C}} = \begin{bmatrix} \bm{I}_9 & \bm{0}_9 & \bm{0}_9 \end{bmatrix}.
\end{equation*}
The other ingredients we need for the Kalman filter are the process
covariance~$\bar{\bm{Q}}$, the measurement covariance~$\bar{\bm{R}}$, and the
initial state covariance~$\bm{P}_0$. In experiment we
use~$\bar{\bm{Q}}=\bar{\bm{B}}\bm{Q}\bar{\bm{B}}^T$ with~$\bm{Q}=10\bm{I}_9$,
$\bar{\bm{R}}=0.001\bm{I}_9$, and~$\bm{P}_0=0.1\bm{I}_{27}$.

\subsubsection{Projectile Kalman Filter}

The discrete-time equations of motion for the projectile are
\begin{equation*}
  \bm{b}^+ = \skew5\bar{\bm{A}}_b\bm{b} + \bar{\bm{B}}_b\bm{g},
\end{equation*}
where
\begin{align*}
  \skew5\bar{\bm{A}}_b &= \begin{bmatrix} \bm{I}_3 & \delta t\bm{I}_3 \\ \bm{0}_3 & \bm{I}_3 \end{bmatrix}, &
  \bar{\bm{B}}_b &= \begin{bmatrix} (1/2)\delta t^2\bm{I}_3 \\ \delta t\bm{I}_3 \end{bmatrix},
\end{align*}
and the measurement model is~$\bm{r}_b=\bar{\bm{C}}_b\bm{b}$, with
\begin{equation*}
  \bar{\bm{C}}_b = \begin{bmatrix} \bm{I}_3 & \bm{0}_3 \end{bmatrix}.
\end{equation*}
Here we use a sampling time of~$\delta t=\SI{10}{ms}$, which is the rate at
which measurements are received from the motion capture system. In experiment
we use process
covariance~$\bar{\bm{Q}}_b=\bar{\bm{B}}_b\bm{Q}_b\bar{\bm{B}}_b^T$
with~$\bm{Q}_b=1000\bm{I}_3$, measurement
covariance~$\bar{\bm{R}}_b=0.001\bm{I}_3$, and initial state
covariance~$\bm{P}_{b_0}=\bm{I}_6$.

\bibliographystyle{IEEEtran}
\bibliography{bibliography}

\end{document}